\pdfoutput=1
\documentclass[11pt]{article}

\usepackage{acl}

\usepackage{times}
\usepackage{latexsym}

\usepackage[T1]{fontenc}
\usepackage[utf8]{inputenc}

\usepackage{microtype}

\usepackage{inconsolata}

\usepackage{multirow}
\usepackage{amsmath}
\usepackage{amssymb}
\usepackage{graphicx}

\definecolor{forestgreen}{rgb}{0.13, 0.55, 0.13}

\title{\textsc{SumTra}: A Differentiable Pipeline\\for Few-Shot Cross-Lingual Summarization}

\author{Jacob Parnell\textsuperscript{1,2}, I\~{n}igo Jauregi Unanue\textsuperscript{1,2}, Massimo Piccardi\textsuperscript{1} \\
 \textsuperscript{1}University of Technology Sydney, Australia \\
 \textsuperscript{2}RoZetta Technology, Australia \\
 \{\texttt{jacob.parnell,inigo.jauregi}\}\texttt{@rozettatechnology.com} \\
 \texttt{massimo.piccardi@uts.edu.au} \\}

\begin{document}
\maketitle
\begin{abstract}
% \textcolor{red}{\textit{(A note for the title: Should we be specific in that we are "revisiting" the pipeline method with a new take? It may make it clearer to a potential reader the issue of novelty discussed among the reviewers in the previous cycle. Perhaps something like "SumTra: Revisiting pipeline-based cross-lingual summarization with soft embeddings" or something that reads better than that?)}}
Cross-lingual summarization (XLS) generates summaries in a language different from that of the input documents (e.g., English to Spanish), allowing speakers of the target language to gain a concise view of their content. In the present day, the predominant approach to this task is to take a performing, pretrained multilingual language model (LM) and fine-tune it for XLS on the language pairs of interest. However, the scarcity of fine-tuning samples makes this approach challenging in some cases. For this reason, 
%inspired by work of \textcolor{red}{(Inigo's TACL citation)}, 
in this paper we propose revisiting the \textit{summarize-and-translate} pipeline, where the summarization and translation tasks are performed in a sequence. This approach allows reusing the many, publicly-available resources for monolingual summarization and translation, obtaining a very competitive zero-shot performance. In addition, the proposed pipeline is completely differentiable end-to-end, allowing it to take advantage of few-shot fine-tuning, where available.
% The tuning step improves the alignment of the models and alleviates error propagation. 
Experiments over two contemporary and widely adopted XLS datasets (CrossSum and WikiLingua) have shown the remarkable zero-shot performance of the proposed approach, and also its strong few-shot performance compared to an equivalent multilingual LM baseline, that the proposed approach has been able to outperform in many languages with only 10\% of the fine-tuning samples.
\end{abstract}

\section{Introduction}
\label{sec:paper4_introduction}
Cross-lingual summarization (XLS) aims to take a document written in a given source language and generate a summary in a chosen target language, providing the speakers of the latter with the ability to concisely understand the content of documents written in foreign languages. However, XLS is a challenging task due to the limited training data which are typically available. Unlike in monolingual summarization, naturally-occurring cross-lingual document-summary pairs are rare, and dedicated XLS human annotation is demanding since it requires uncommon skills of the annotators \cite{wang-etal-2022-survey}. This has often led to the reuse of existing multilingual data with post-hoc alignments for cross-lingual use \citep{ladhak-etal-2020-wikilingua, bhattacharjee-etal-2022-crosssum}.

Given the constraints in dedicated training resources, most recent approaches have focused on employing existing multilingual LMs \citep{liu-etal-2020-multilingual-denoising, tang-etal-2021-multilingual, xue-etal-2021-mt5}, pretrained in the typical unsupervised manner over large corpora, and fine-tuning them with the limited XLS resources available for the chosen language pairs \citep{perez-beltrachini-lapata-2021-models, ma-etal-2021-deltalm}. However, these multilingual models suffer from well-known limitations. On the one hand, the uneven pretraining of multilingual LMs across languages often results in poor knowledge transfer to low-resource languages \citep{joshi-etal-2020-state, bhattacharjee-etal-2022-crosssum}. On the other hand, the superposition of too many languages in a single model can result in a degradation of cross-lingual performance in the downstream task (i.e., language interference) \citep{pfeiffer-etal-2022-lifting}. In addition, it is not trivial to reuse the abundant, existing monolingual summarization data, since fine-tuning a multilingual LM with monolingual data often compromises its ability to generate text in a language different from the input's \citep{vu-etal-2022-overcoming, bhattacharjee-etal-2022-crosssum}---a problem known as ``catastrophic forgetting'' \citep{vandeven-2019-scenarios}. The above issues compound in the impossibility of achieving a satisfactory zero-shot and few-shot XLS performance out of conventional multilingual LMs.

For this reason, this work revisits the \textit{summarize-and-translate} approach to XLS~\citep{wan-etal-2010-cross}, with the main aim of fully leveraging the existing monolingual summarization resources (i.e., training data, pretrained models) to obtain a performing zero-shot XLS pipeline. Specifically, we propose combining 1) a monolingual summarizer trained with abundant resources in the source language with 2) a pretrained machine translation model that translates into the target language. If the quality of both models is high, such a pipeline should be able to achieve a significant zero-shot performance. Yet, it can also suffer from model misalignment and error propagation. Therefore, we modify the summarizer to output ``soft'' predictions, ensuring that the pipeline remains fully differentiable end-to-end \cite{subramanian-etal-2017-adversarial, kumar-etal-2021-controlled, unanue-etal-2023-t3l}. This allows fine-tuning it to improve the coupling of the models, alleviate error propagation, and obtain summaries that are closer to the ideal, joint summarization/translation of the XLS task. For immediacy, we refer to the proposed pipeline as \textsc{SumTra}.

In particular, in this paper we focus on the less explored \textit{English-to-many} XLS task (most work to date has focused on many-to-English \citep{zhu-etal-2019-NCLS, ladhak-etal-2020-wikilingua, ma-etal-2021-deltalm, chi-etal-2021-mt6} or specific language pairs such as English-to-Chinese \citep{ayana-etal-2018-zero, zhu-etal-2019-NCLS, bai-etal-2021-cross, liang-etal-2022-variational}. We believe that this is a valuable contribution as it provides access to summaries of the multitude of existing English documents for speakers of other languages around the world. To this aim, we have carried out experiments over two widely used XLS datasets (CrossSum \citep{bhattacharjee-etal-2022-crosssum} and WikiLingua \citep{ladhak-etal-2020-wikilingua}), with a range of language pairs spanning high-, medium-, and low-resource languages. The results show a strong quantitative performance for the zero-shot pipeline, and a competitive edge over comparable multilingual language model baselines with up to 1000-shot fine-tuning\footnote{Our code is publicly accessible at: \url{https://github.com/jacob-parnell-rozetta/sumtra}}.

Overall, our paper makes the following contributions:
\begin{itemize}
    \item A \textit{summarize-and-translate} pipeline that leverages contemporary state-of-the-art language models (and their resources) for the summarization and  translation steps.

    \item  A fully differentiable approach through the use of ``soft'' summaries, making the pipeline fine-tunable end-to-end.
    
    \item A novel objective function that incorporates a back-translation loss over the summarization module to ground the generation of the intermediate summaries to the target language reference.
    
    \item A comparative experimental evaluation of the proposed approach over two popular cross-lingual summarization datasets spanning two diverse domains, including an extensive qualitative, ablation, and sensitivity analysis.

\end{itemize}

\section{Related Work}
\label{sec:paper4_related_work}

Cross-lingual summarization (XLS) has been an active research topic for a long time \citep{leuski-etal-2003-custard, wan-etal-2010-cross}. Pre-neural methods have often combined monolingual summarization and machine translation (MT) modules into pipeline approaches that \textit{summarize-and-translate} \citep{orasan-chiorean-2008-evaluation, wan-etal-2010-cross}, or \textit{translate-and-summarize} \citep{leuski-etal-2003-custard, wan-2011-using, boudin-etal-2011-graph}. While conceptually justifiable, these approaches inevitably suffered from error propagation between the modules, and, obviously, the architectural limitations of the models of the day \citep{zhu-etal-2019-NCLS, ouyang-etal-2019-robust} . 

With the recent development of multilingual pretrained language models such as mBART \citep{lewis-etal-2020-bart} and mT5 \citep{xue-etal-2021-mt5}, there has been a surge in XLS research that has focused on fine-tuning these models with XLS datasets, and as a consequence has relegated pipeline methods to be regarded as mere baselines for comparison \citep{ladhak-etal-2020-wikilingua, dou-etal-2020-deep, perez-beltrachini-lapata-2021-models}. However, the current approaches are not exempt from performance limitations at their turn, in particular when applied to low-resource languages\footnote{We note that in the XLS task there are many dimensions in which a language can be ``low-resource'', namely: the monolingual data for model pretraining; the parallel corpora for translation pretraining; and the annotated XLS document-summary pairs for fine-tuning.}. To address them, \citet{bhattacharjee-etal-2022-crosssum} has attempted to transfer knowledge from high- to low-resource languages by a multi-stage sampling algorithm that aptly up-samples the low-resource languages. Other works have explored using language-specific adapter modules in various cross-lingual tasks \citep{rebuffi-etal-2017-learning, houlsby-etal-2019-parameter-efficient} to increase the linguistic capacity of the model at a parity of trainable parameters and alleviate language interference \citep{pfeiffer-etal-2022-lifting}. \citet{bai-etal-2021-cross} have proposed using a combination of monolingual and cross-lingual summarization in an attempt to improve performance on low-resource languages.  More recently, \citet{wang-etal-2023-towards-unifying} has proposed leveraging various large ($>$100B parameters) language models for zero-shot cross-lingual summarization. By contrast,  in this paper we intentionally focus on the utilization of much smaller, modular, and trainable models in the zero- and few-shot scenario.

\section{SumTra}
\label{sec:paper4_proposed_method}
The proposed \textsc{SumTra} model consists of the cascade of two language models: a monolingual summarization language model, followed by a machine translation language model, which we refer to as \textsc{Sum} and \textsc{Tra} for \textit{summarize} and \textit{translate}, respectively.

Let us denote the token sequence of the input document as $x = \{x_1, \ldots x_n\}$, and the token predicted by the \textsc{Sum} module at slot $j$ as $s_j$. We can then express the sequence of probability vectors output by the \textsc{Sum} module over the vocabulary as $\{\textbf{p}_1,...\textbf{p}_j...,\textbf{p}_m\}$, with:

\begin{equation}
\label{equation:paper4_mslm_vocab}
    \textbf{p}_j = \text{\textsc{Sum}}(s_{j-1},x,\theta)
\end{equation}

\noindent where $s_{j-1}$ is the previous predicted token and $\theta$ are the module's parameters. For simplicity and efficiency we use greedy search for token prediction, but in principle any decoding approach can be used.

The probability vectors $\{\textbf{p}_1,...\textbf{p}_j...,\textbf{p}_m\}$ are then individually mixed with the embedding layer $\textbf{E}$ of the \textsc{Tra} module of size $D \times V$ (embedding $\times$ vocabulary) to obtain a sequence of expected embeddings, $\textbf{e} = \{\textbf{e}_1...\textbf{e}_j...\textbf{e}_m\}$, with:

\begin{equation}
\label{equation:paper4_expected_embs}
    \textbf{e}_j = \mathbb{E}[\textbf{E}]_{\textbf{p}_j} = \textbf{E} \ \textbf{p}_j
\end{equation}

\noindent which are equivalent to ``soft'' predictions from the \textsc{Sum} module. These expected embeddings, which represent the intermediate summary, are then provided as input to the \textsc{Tra} module bypassing its embedding layer. Eventually, the \textsc{Tra} module predicts the translation in the target language:

\begin{equation}
\label{equation:paper4_translation_prediction}
    \bar{y} = \text{\textsc{Tra}}(\textbf{e}, \sigma)
\end{equation}

\noindent where $\bar{y}$ denotes the translation and $\sigma$ the module's parameters. Since the soft predictions from the \textsc{Sum} module do not interrupt backpropagation, the whole network can be trained end-to-end.

For fine-tuning the entire \textsc{SumTra} model, we use the standard negative log-likelihood:
\begin{equation}
\label{equation:paper4_main_loss}
    \textsc{NLL} = -\sum\limits_{t=1}^{T}\log p(y_{t}|y_{1}, \dots y_{t-1}, \textbf{e}, \theta, \sigma) \\
\end{equation}

\noindent where with $\{y_{1}, \dots y_{T}\}$ we denote the sequence of ground-truth tokens in the target language, and with $p(y)$ the probabilities output by the translator. 

However, fine-tuning the \textsc{Sum} module with only the standard negative log-likelihood of the ground-truth summary in the target language allows for too many degrees of freedom in the generation of the intermediate English summary, and can lead to inaccurate summaries with respect to the source document. For this reason, we add an auxiliary training objective that encourages the predicted summary to adhere to the target more closely. To this aim, we first back-translate the ground-truth sequence, $y$, into the language of the summarizer (i.e., English) using a reverse \textsc{Tra} module, and then use it as auxiliary training objective for the summarizer:
\begin{equation}
\label{equation:paper4_auxiliary_loss}
    \textsc{NLL}_{\textsc{Sum}} = -\sum\limits_{t=1}^{T}\log p(\hat{y}_{t}|\hat{y}_{1}, \dots \hat{y}_{t-1}, x, \theta) \\
\end{equation}

\noindent where $\hat{y}$ denotes the back-translated sequence, and $p(\hat{y})$ the probabilities output by the summarizer. We note  that our use of a separate summarization module would also allow using other typical summarization training objectives such as  sentence-level coherence \citep{li-etal-2019-deep}, coverage of the input document \citep{parnell-etal-2022-multi} and so forth, but we have decided to leave this exploration to future work.

The training objectives in Equations \ref{equation:paper4_main_loss} and \ref{equation:paper4_auxiliary_loss}, are eventually combined in a simple convex combination:

\begin{equation}
\label{eq:mixed_objective}
    L = \alpha \textsc{NLL}_{\textsc{Sum}} + (1 - \alpha) \textsc{NLL} \\
\end{equation}

\noindent using a scaling coefficient, $\alpha$, that acts as a hyperparameter in the loss. We have set $\alpha$ to $0.99$ for all experiments, and report a sensitivity analysis in Appendix \ref{subsec:paper4_appendix_alpha_weights}. 

\section{Experimental Setup}
\label{sec:paper4_experiments}
\subsection{Datasets, Baselines, Evaluation Metrics}
\label{subsec:paper4_datasets_baselines_evalmetrics}
We have carried out extensive zero and few-shot experiments over twelve English-to-many language pairs from the CrossSum \citep{bhattacharjee-etal-2022-crosssum} and WikiLingua \citep{ladhak-etal-2020-wikilingua} datasets. We have selected six languages from each dataset, and categorized them as high-, medium- and low-resource based on the number of sentences used for the pretraining of the respective language in our main baseline, mBART-50 \citep{tang-etal-2021-multilingual}.

To implement the proposed approach, we have used the mBART-50 one-to-many\footnote{\url{https://huggingface.co/facebook/mbart-large-50-one-to-many-mmt}} variant for the \textsc{Tra} module, and the many-to-one\footnote{\url{https://huggingface.co/facebook/mbart-large-50-many-to-one-mmt}} variant for both the \textsc{Sum} module and the generation of the back-translations used for fine-tuning (Equation \ref{equation:paper4_auxiliary_loss}). The back-translations have been generated once and for all offline, and added to the dataset.

As baselines, we have employed various, strong multilingual models that include: 1) the mT5-m2m model of \citet{bhattacharjee-etal-2022-crosssum}, fine-tuned on all languages and full training splits of the CrossSum dataset; 2) a pretrained mBART-50 \citep{tang-etal-2021-multilingual}, both with and without an initial training with a monolingual English dataset (respectively, mBART-50-mono and mBART-50 in the following); 3) two large language models from Open AI (ChatGPT and davinci-003), leveraging a ``direct'' and ``summarize-then-translate'' prompt, respectively, as defined in \citet{wang-etal-2023-zeroshot}, and 4) the \textsc{Pisces} model of \citet{wang-etal-2023-towards-unifying} -- a modified mBART-50 model that leverages extra cross-lingual and task-specific pretraining over huge resources (20.6M samples from the OPUS parallel corpora and 3.1 from mC4, respectively).

To evaluate the predictions, we have used ROUGE \citep{lin-2004-rouge} and its multilingual adaptation\footnote{For brevity, we will refer to ``ROUGE'' as ``mROUGE'' throughout, to accommodate all languages. Details on mROUGE are provided in Appendix \ref{subsec:paper4_appendix_expsetup}.}, mROUGE \citep{conneau-lample-2019-crosslingual}, which leverages language-specific tokenizers and stemmers  to pre-process non-English text prior to a standard ROUGE calculation. We have computed the ROUGE scores as an average of ROUGE-1, ROUGE-2 and ROUGE-L F1. Similarly to \citet{koto-etal-2021-evaluating}, we also report BERTScore \citep{zhang-etal-2020-bertscore} for its ability to better assess the semantic alignment of the predictions and the references.

\subsection{Model Training}
\label{subsec:paper4_model_training}
Prior to running the XLS experiments, we have trained the \textsc{Sum} module for monolingual summarization in  English. To this aim, we have leveraged the respective English-English training split of CrossSum or WikiLingua\footnote{Appendix \ref{subsec:paper4_appendix_other_monolingual} explores other options for the monolingual summarization training.}, and chosen the best performing checkpoint based on a validation criterion. For the experiments in the few-shot fine-tuning configuration, we have chosen to fine-tune the entire \textsc{SumTra} model; however, it is also possible to freeze either the summarization or the translation module, and we present an ablation in Section  \ref{subsec:paper4_specific_module_tuning}. Further details of the experimental setup are provided in Appendixes \ref{subsec:paper4_appendix_expsetup} and \ref{subsec:paper4_appendix_modelhyp}.

\section{Results and Analysis}
\label{sec:paper4_results}

% Table 1
\begin{table*}[!ht]
\begin{center}
\resizebox{0.99\textwidth}{!}{%
\begin{tabular}{l|c|c|c|c|c|c|c}
\hline
\multirow{2}{*}{\textbf{Model}} &\multicolumn{2}{c|}{High} & \multicolumn{2}{c|}{Medium} & \multicolumn{2}{c|}{Low} & \multirow{2}{*}{\textbf{Average}} \\ \cline{2-7}
& en-es$^{\dagger}$ & en-fr$^{\dagger}$ & en-ar$^{\dagger}$ & en-uk & en-az & en-bn$^{\dagger}$ \\
\hline
mBART-50 (fully fine-tuned) & 21.04 / 55.98 & 17.18 / 51.18 & 18.14 / 61.87 & 7.62 / 59.34 & 13.98 / 54.53 & 7.58 / 61.41 & 14.25 / 57.39 \\
\hline
mBART-50 (0-shot) & 1.18 / 26.46 & 0.26 / 21.14 & 0.85 / 33.62 & 0.00 / 28.96 & 0.11 / 19.79 & 0.00 / 25.83 & 0.40 / 25.97 \\
% mBART-50 (10-shot) & 1.17 / 26.55 & 0.26 / 21.11 & 1.10 / 35.28 & 0.00 / 28.96 & 0.17 / 20.66 & 0.00 / 25.83  &  0.45 / 26.40 \\
mBART-50 (50-shot) & 1.18 / 26.54 & 0.26 / 21.06 & 1.27 / 36.14 & 0.00 / 28.96 & 0.17 / 20.56 & 0.00 / 25.00 & 0.48 / 26.38 \\
mBART-50 (100-shot) &  1.18 / 26.50 & 14.53 / 48.42 & 1.28 / 36.20 & 4.46 / 54.69 & 0.17 / 20.57 & 0.81 / 39.70 &  3.74 / 37.68 \\
mBART-50 (1000-shot) & 18.29 / 53.99 & 17.57 / 50.76 & 14.36 / 60.06 & 7.41 / 58.01 & 14.32 / 54.74 & 7.17 / 60.53 & 13.19 / 56.35 \\
\hline
mBART-50-mono (0-shot) & 5.39 / 29.98 & 4.97 / 31.58 & 0.20 / 21.74 & 1.75 / 23.47 & 2.00 / 21.84 & 0.00 / 16.31 & 2.39 / 24.15 \\
mBART-50-mono (50-shot) & 5.42 / 30.11 & 4.98 / 31.60 & 0.20 / 21.74 & 1.78 / 23.48 & 1.99 / 22.01 & 0.00 / 16.33 & 2.40 / 24.21 \\
mBART-50-mono (100-shot) & 5.66 / 30.76 & 4.88 / 31.64 & 0.20 / 21.73 & 1.73 / 23.56 & 2.18 / 21.59 & 0.00 / 16.42 & 2.44 / 24.28 \\
mBART-50-mono (1000-shot) & 18.65 / 54.06 & 16.69 / 50.91 & 12.52 / 58.38 & 7.52 / 56.67 & 13.56 / 51.68 & 7.78 / 62.62 & 12.79 / 55.72 \\
\hline
\textsc{SumTra} (0-shot) & 20.19 / 55.41 & 20.87 / 53.98 & 15.80 / 60.33 & 8.74 / 59.80 & 13.28 / 54.09 & 4.04 / 54.32 & 13.82 / 56.32 \\
% \textsc{SumTra} (10-shot) & 20.46 / 55.99 & 20.97 / \textbf{54.20} & 15.67 / 60.34 & 9.05 / 59.67 & 13.89 / 54.34 & 3.62 / 54.43 & 13.94 / 56.50 \\
\textsc{SumTra} (50-shot) & 21.32 / 56.66 & 20.03 / 53.46 & 15.84 / 60.62 & 8.76 / 59.88 & 14.68 / 54.54 & 3.90 / 54.85 & 14.09 / 56.67 \\
\textsc{SumTra} (100-shot) & 21.47 / 56.41 & \textbf{21.24} / \textbf{54.06} & 16.08 / 60.67 & 9.47 / 59.98 & 13.97 / 54.10 & 4.67 / 56.28 & 14.47 / 56.92 \\
\textsc{SumTra} (1000-shot) & 21.29 / 56.41 & 20.30 / 53.94 & \textbf{17.57} / \textbf{61.73} & \textbf{10.17} / \textbf{60.48} & 15.74 / 55.94 & 6.11 / 58.58 & 15.20 / 57.85 \\
\hline
\hline
% $\textsc{MLE}_{\textsc{XLS+MT+DIS}}$ \citep{dou-etal-2020-deep} \textbf{*} & / & / & / & / & / & / & / \\
mT5-m2m \citep{bhattacharjee-etal-2022-crosssum} & \textbf{22.23} / \textbf{56.86} & 19.27 / 52.48 & 16.56 / 60.49 & 8.63 / 59.65 & \textbf{18.48} / \textbf{57.27} & \textbf{11.49} / \textbf{66.31} & \textbf{16.11} / \textbf{58.84} \\
davinci-003 (ST) \citep{wang-etal-2023-zeroshot} & 13.71 / 50.74 & 6.58 / 24.46 & 8.74 / 55.60 & 5.52 / 54.96 & 9.17 / 49.27 & 4.82 / 61.66 & 8.09 / 49.45 \\
ChatGPT (Direct) \citep{wang-etal-2023-zeroshot} & 16.20 / 52.02 & 13.75 / 47.41 & 10.24 / 56.36 & 4.03 / 54.78 & 11.14 / 47.85 & 3.99 / 60.69 & 9.89 / 53.19 \\
\textsc{Pisces} \citep{wang-etal-2023-towards-unifying} & 3.02 / 31.92 & 9.93 / 42.73 & 0.08 / 44.65 & 0.73 / 39.56 & 3.04 / 35.77 & 0.00 / 53.63 & 2.80 / 41.38 \\
% \textsc{Pisces} (50-shot) \citep{wang-etal-2023-towards-unifying} & 10.85 / 45.57 & 9.83 / 42.74 & 0.07 / 44.42 & 0.73 /39.20 & 2.89 / 40.58 & 0.00 / 53.57 & 4.06 / 44.35 \\
% \textsc{Pisces} (100-shot) \citep{wang-etal-2023-towards-unifying} & 14.54 / 50.79 & 9.87 / 42.74 & 0.06 / 44.28 & 0.87 / 38.43 & 2.53 / 40.00 & 0.00 / 52.36 & 4.65 / 44.77 \\
%\textsc{Pisces} (1000-shot) \citep{wang-etal-2023-towards-unifying} & 18.57 / 53.99 & 17.23 / 50.57 & 14.10 / 59.69 & 6.14/ 57.27 & 15.59 / 55.36 & 7.86 / 62.92 & 13.25 / 56.63 \\
% \hline
\hline
\end{tabular}
}
\caption{Results for the CrossSum dataset, grouped into high, medium, and low-resource languages. We report the average of ROUGE-1, ROUGE-2, and ROUGE-L F1 (or the mROUGE equivalent where applicable as denoted with $\dagger$) and BERTScore. The best scores are boldfaced.}
\label{tab:paper4_results_crosssum}
\end{center}
\end{table*}

% Table 2 
\begin{table*}[!ht]
\begin{center}
\resizebox{0.99\textwidth}{!}{%
\begin{tabular}{l|c|c|c|c|c|c|c}
\hline
\multirow{2}{*}{\textbf{Model}} &\multicolumn{2}{c|}{High} & \multicolumn{2}{c|}{Medium} & \multicolumn{2}{c|}{Low} & \multirow{2}{*}{\textbf{Average}} \\ \cline{2-7}
& en-ru$^{\dagger}$ & en-zh$^{\dagger}$ & en-ar$^{\dagger}$ & en-tr$^{\dagger}$ & en-th$^{\dagger}$ & en-id & \\
\hline
mBART-50 (fully fine-tuned) & 17.10 / 62.09 & 24.01 / 65.21 & 16.95 / 65.17 & 18.56 / 59.71 & 26.77 / 70.94 & 19.28 / 60.29 & 19.73 / 63.90 \\
\hline
mBART-50 (0-shot) & 0.57 / 29.54 & 0.00 / 36.75 & 0.78 / 33.29 & 0.91 / 23.08 & 1.78 / 31.11 & 0.94 / 26.44 & 0.83 / 30.04 \\
% mBART-50 (10-shot) & 0.63 / 29.93 & 0.00 / 36.75 & 0.79 / 33.60 & 0.91 / 23.08 & 1.78 / 31.11 & 1.05 / 27.06 & 0.86 / 30.26 \\
mBART-50 (50-shot) & 0.71 / 30.69 & 0.00 / 36.75 & 0.78 / 34.19 & 1.02 / 23.56 & 1.71 / 31.04 & 1.25 / 27.54 & 0.91 / 30.63 \\
mBART-50 (100-shot) & 6.77 / 52.70 & 0.00 / 36.75 & 0.79 / 34.09 & 6.70 / 47.84 & 0.63 / 31.77 & 1.25 / 27.32 & 2.69 / 38.41 \\
mBART-50 (1000-shot) & 9.43 / 56.49 & 20.35 / 62.06 & 11.11 / 61.74 & 15.08 / 56.74 & 19.65 / 61.71 & 10.95 / 53.01 & 14.43 / 58.63 \\
\hline
mBART-50-mono (0-shot) & 0.58 / 31.85 & 9.01 / 36.00 & 0.28 / 26.03 & 2.24 / 28.79 & 12.79 / 29.02 & 2.06 / 32.35 & 4.49 / 30.67 \\
mBART-50-mono (50-shot) & 0.57 / 31.86 & 8.98 / 36.00 & 0.28 / 26.03 & 2.24 / 28.78 & 12.79 / 29.02 & 2.05 / 32.34 & 4.48 / 30.68 \\
mBART-50-mono (100-shot) & 0.58 / 31.85 & 8.98 / 36.02 & 0.28 / 26.03 & 2.24 / 28.78 & 12.79 / 29.03 & 2.05 / 32.34 & 4.49 / 30.68 \\
mBART-50-mono (1000-shot) & 11.16 / 58.41 & 20.36 / 62.37 & 10.09 / 60.41 & 13.69 / 54.74 & 22.25 / 67.32 & 11.57 / 53.36 & 14.85 / 59.44 \\
\hline
\textsc{SumTra} (0-shot) & 10.35 / 56.12 & 21.13 / 57.24 & 11.61 / 61.48 & 10.96 / 53.96 & 14.66 / 51.39 & 12.83 / 54.84 & 13.59 / 55.84 \\
% \textsc{SumTra} (10-shot) & 10.23 / 56.14 & 20.26 / 55.74 & 10.96 / 61.07 & 10.24 / 53.10 & 14.93 / 51.51 & 11.52 / 53.26 & 13.02 / 55.14 \\
\textsc{SumTra} (50-shot) & 11.73 / 58.33 & 19.70 / 60.16 & 11.74 / 61.79 & 11.44 / 54.78 & 15.83 / 53.04 & 12.79 / 55.06 & 13.87 / 57.19 \\
\textsc{SumTra} (100-shot) & 12.01 / 58.85 & 19.70 / 61.08 & 11.58 / 61.66 & 12.50 / 55.69 & 16.15 / 54.16 & 13.12 / 55.68 & 14.18 / 57.85 \\
\textsc{SumTra} (1000-shot) & \textbf{13.38} / \textbf{59.85} & 21.13 / 63.12 & \textbf{13.04} / \textbf{62.61} & \textbf{16.23} / \textbf{57.94} & 18.93 / 58.87 & \textbf{14.67} / \textbf{57.09} & \textbf{16.23} / \textbf{59.91} \\
\hline
\hline
% $\textsc{MLE}_{\textsc{XLS+MT+DIS}}$ \citep{dou-etal-2020-deep} \textbf{*} & / & / & / & / & / & / & / \\
davinci-003 (ST) \citep{wang-etal-2023-zeroshot} & 10.37 / 53.19 & 10.80 / 38.48 & 8.78 / 56.23 & 
9.55 / 52.25 & 12.84 / 58.84 & 10.37 / 50.45 & 10.45 / 51.57 \\
ChatGPT (Direct) \citep{wang-etal-2023-zeroshot} & 8.52 / 52.55 & 15.33 / 53.19 & 7.34 / 55.18 & 9.24 / 53.17 & 10.45 / 58.07 & 10.75 / 51.30  & 10.27 / 53.91 \\
\textsc{Pisces} \citep{wang-etal-2023-towards-unifying} & 0.59 / 34.25 & \textit{\textbf{42.65}} / \textit{\textbf{73.66}} & 0.34 / 41.99 & 4.32 / 38.73 & \textit{\textbf{47.13}} / \textit{\textbf{78.60}} & 1.83 / 43.21 & 16.14 / 51.74 \\
% \textsc{Pisces} (50-shot) \citep{wang-etal-2023-towards-unifying} & 0.55 / 34.95 & 45.24 / 74.87 & 0.34 / 41.84 & 8.44 / 49.63 & 48.85 / 79.32 & 1.82 / 43.08 & 17.54 / 53.95 \\
% \textsc{Pisces} (100-shot) \citep{wang-etal-2023-towards-unifying} & 0.55 / 34.74 & 45.73 / 75.19 & 0.34 / 41.30 & 12.11 / 53.68 & 48.65 / 79.21 & 1.84 / 42.73 & 18.20 / 54.48 \\
%\textsc{Pisces} (1000-shot) \citep{wang-etal-2023-towards-unifying} & 13.01 / \textbf{59.85} & \textbf{46.16} / \textbf{75.38} & \textbf{13.32} / \textbf{62.90} & \textbf{21.35} / \textbf{61.56} & \textbf{48.37} / \textbf{79.78} &  \textbf{16.16} / \textbf{58.27} & \textbf{26.39} / \textbf{66.29} \\
% \hline
\hline
\end{tabular}
}
\caption{Results for the WikiLingua dataset, grouped into high, medium, and low-resource languages. We report the average of ROUGE-1, ROUGE-2, and ROUGE-L F1 (or the mROUGE equivalent where applicable as denoted with $\dagger$) and BERTScore. The best scores are boldfaced. The italicized results are commented upon in Section \ref{sec:paper4_results}.}
\label{tab:paper4_results_wiki}
\end{center}
\end{table*}

Tables \ref{tab:paper4_results_crosssum} and \ref{tab:paper4_results_wiki} present the results of the proposed approach and comparative baselines over the chosen language pairs, grouped into high-, medium-, and low-resource languages, for the CrossSum and WikiLingua datasets, respectively.

\textbf{\textsc{SumTra} vs. mBART-50.} In both tables, we compare the proposed \textsc{SumTra} model with both mBART-50 and mBART-50-mono (the version with an initial English summarization training), and in both zero- and few-shot configurations (50-1000 examples). The results show that the English training can be beneficial for improving the average zero- and few-shot performance of mBART-50 \citep{wang-etal-2022-clidsum}; however, the results are not consistent across languages, even for those that are linguistically similar (e.g., Spanish and French). \textsc{SumTra} comparatively displays much stronger average zero- and few-shot performance up to and including 1000 shots, showing the usefulness of the proposed approach. For instance, \textsc{SumTra} (0-shot) outperforms both mBART-50 variants with 1000 shots  on average over the CrossSum languages. In a similar fashion, at a parity of fine-tuning samples (1000-shots), the most performant \textsc{SumTra} model outperforms mBART-50 by +1.28 BERTScore pp on average over the WikiLingua languages.

\textbf{\textsc{SumTra} vs. \textsc{Pisces}.} We also compare \textsc{SumTra} against \textsc{Pisces}, but for brevity, limit the experiments to the zero-shot configuration downloaded from \url{https://huggingface.co/Krystalan/PISCES}. The results show a comparatively rather modest performance from \textsc{Pisces}, with the exception of two staggering results for the Chinese and Thai languages of WikiLingua. Since these scores are much higher than those reported in \citet{wang-etal-2023-towards-unifying} for a fully fine-tuned \textsc{Pisces} model, we speculate that there may exist some overlap between some of their training data and our test sets. An alternative explanation is that Chinese and Thai were part of \textsc{Pisces}' pre-training languages, and the alignment with their WikiLingua's test sets may have proved extraordinarily effective. For all other languages, \textsc{SumTra} has displayed a much stronger zero-shot performance compared to \textsc{Pisces}, confirming the validity of our approach.

\textbf{\textsc{SumTra} vs. mT5/ChatGPT/davinci-003.} Lastly, we compare \textsc{SumTra} to the remaining baselines: the mT5 many-to-many model, ChatGPT, and davinci-003. We note that the mT5 model has been fine-tuned over all the language pairs in the CrossSum dataset (1,500+), and with the entire available XLS training set ($\sim$900-1,500 samples per language pair) \citep{bhattacharjee-etal-2022-crosssum}, and should therefore be regarded in Table \ref{tab:paper4_results_crosssum} as a hard-to-near upper bound. With that said, \textsc{SumTra} has obtained higher scores for 3 of the 6 languages, and competitive scores for the other three. Lastly, ChatGPT and davinci-003 have obtained some of the lowest average mROUGE and BERTScore scores compared to the other models, showing that they lack the task-specific capability that even a few-shot mBART-50 or \textsc{SumTra} model displays.

Overall, these results  show that the proposed \textsc{SumTra} model is capable of a very strong zero-shot performance, and with a few-shot fine-tuning can reach or near state-of-the-art performance. This can prove particularly useful for languages with a scarcity ($\le$ 100) of annotated XLS samples.

\subsection{Alternative Monolingual Training}
\label{subsec:paper4_appendix_other_monolingual}
\begin{figure}[!t]
    % \centering
    \includegraphics[width=.95\linewidth, trim={12 10 0 10}, clip]{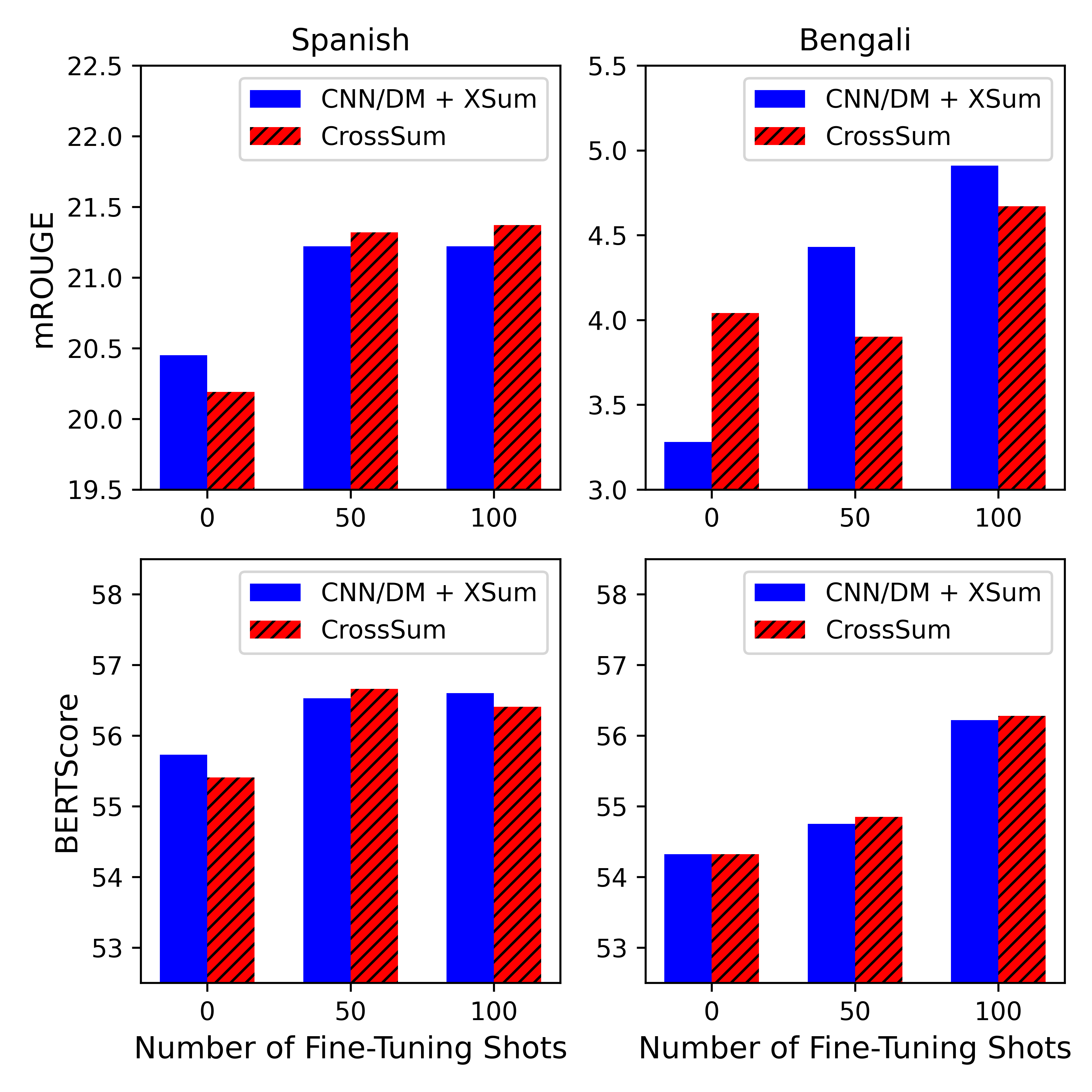}
    \caption{Performance comparison between \textsc{SumTra} models trained with CNN/DM and XSum, and with the CrossSum English training split.}
    \label{fig:paper4_appendix_other_monolingual}
\end{figure}

Given the vast amounts of available English summarization datasets, we have also explored training the \textsc{Sum} module with two widespread datasets, CNN/DailyMail \citep{see-etal-2017-get} and XSum \citep{narayan-etal-2018-dont} in alternative to the English training splits of the XLS datasets. For simplicity, we have first trained the summarizer on CNN/DM, and then continued training on XSum. We have then performed zero-, 50-, and 100-shot fine-tuning of \textsc{SumTra}, and compared the performance with the model trained on the CrossSum English split. The results over the Spanish and Bengali test sets are displayed in Figure \ref{fig:paper4_appendix_other_monolingual}, showing that the performance has been approximately on par and always close. We can then argue that re-training the summarizer for every specific XLS dataset may be unnecessary, and that the zero-shot performance of the proposed approach trained with generic English summarization resources is likely to remain competitive over a variety of domains.

\subsection{Cross-Domain Analysis}
\label{subsec:paper4_appendix_cross_domain_exps}
\begin{figure}[!t]
    % \centering
    \includegraphics[width=.95\linewidth, trim={0 10 10 10}, clip]{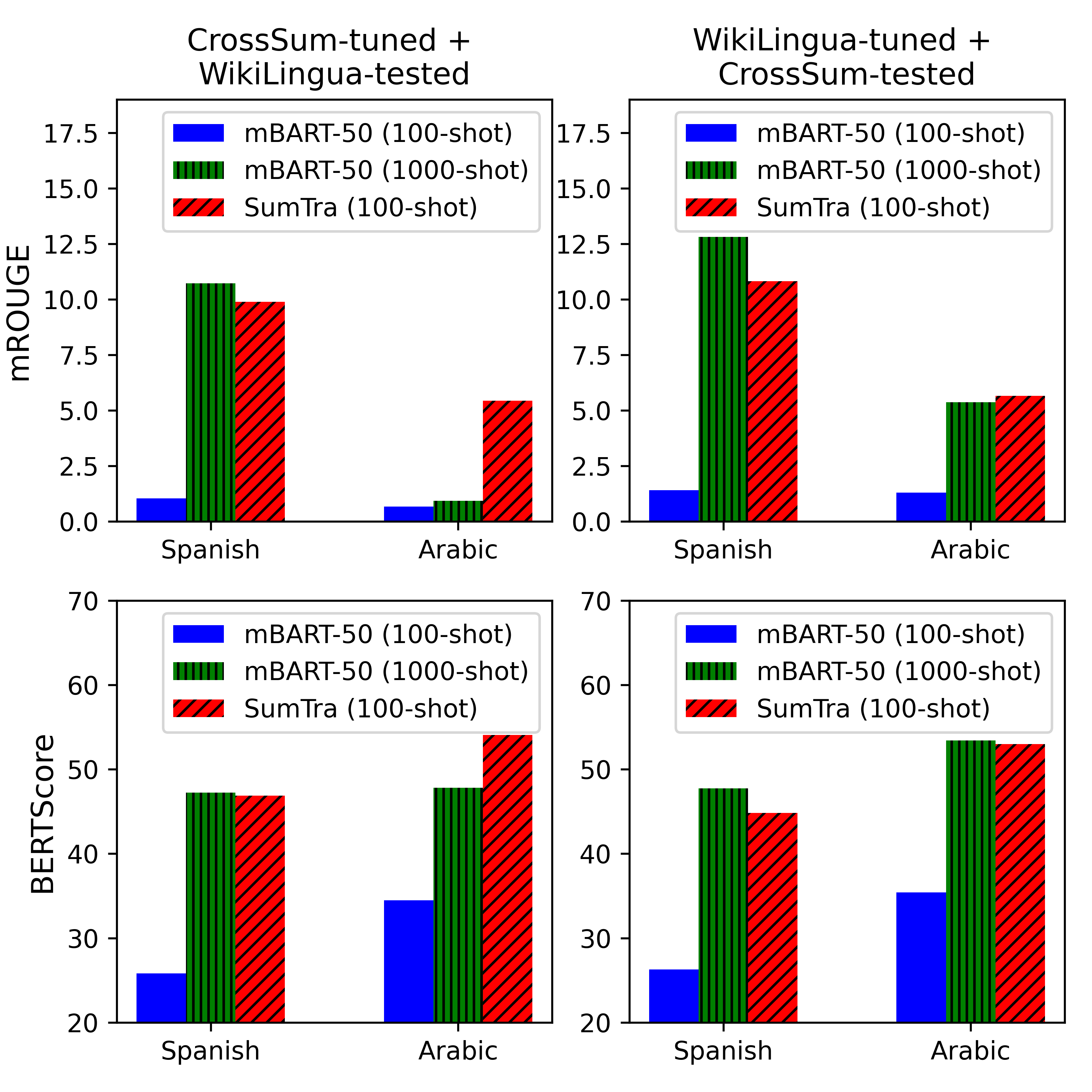}
    \caption{Cross-domain mROUGE/BERTScore scores for Spanish and Arabic. Left: CrossSum-tuned and WikiLingua-tested; Right: vice versa. We have also included mBART-50 (1000-shot) to highlight \textsc{SumTra}'s few-shot capability.}
    \label{fig:paper4_appendix_results_crossdomain}
\end{figure}

In addition, we have explored the cross-domain robustness of \textsc{SumTra} by training and fine-tuning the model on one dataset and testing it on the other (i.e., training with CrossSum and testing on WikiLingua, and vice versa). Figure \ref{fig:paper4_appendix_results_crossdomain} shows the results for \textsc{SumTra} and an equivalent mBART-50 model, both fine-tuned with 100-shots in Spanish and Arabic from one dataset, and tested in the same language on the other. We also report the results for mBART-50 fine-tuned with 1000 shots to show the competitiveness of our approach with just 10\% of the fine-tuning samples.

Overall, the result trends shown in Figure \ref{fig:paper4_appendix_results_crossdomain} are significantly lower than those in Tables \ref{tab:paper4_results_crosssum}
and \ref{tab:paper4_results_wiki}; however, the performance gap between \textsc{SumTra} (100-shot) and mBART-50 (100-shot) has remained wide. These results further highlight the benefits of the proposed pipeline-based approach, as they show that it generalizes reasonably well across domains (news for CrossSum and how-to articles for WikiLingua), particularly in a few-shot setting. mBART-50 (1000-shot) has been able to marginally outperform \textsc{SumTra} (100-shot) in some cases.

\subsection{The Catastrophic Forgetting Problem}
\label{subsec:paper4_catastrophic_forgetting}

In the context of multilingual models, the catastrophic forgetting problem refers to the drop in multilingual performance for models that have been trained with monolingual task data  \citep{pfeiffer-etal-2022-lifting}. \citet{bhattacharjee-etal-2022-crosssum} have explored this within their mT5-m2m model and shown that its zero-shot cross-lingual performance is very poor despite its extensive multilingual pretraining with a multitude of language pairs. Therefore, in this section we set to explore how catastrophic forgetting behaves in the XLS case within a zero-shot, few-shot and full fine-tuning scenarios.

\begin{figure}[!t]
    \centering
    \includegraphics[width=0.99\linewidth, trim={0 0 0 0cm}, clip]{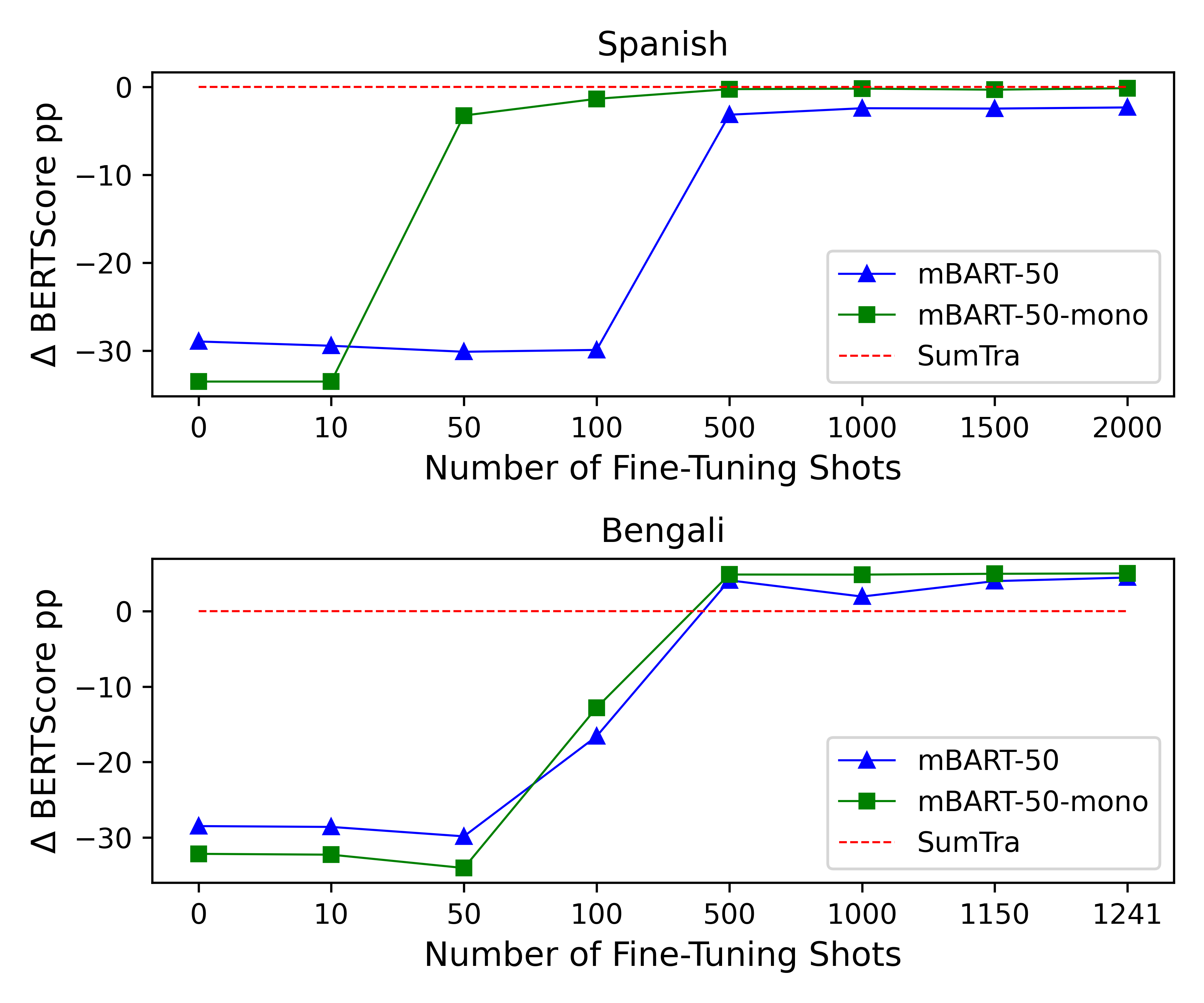}
    \caption{Exploring the catastrophic forgetting problem with mBART-50, mBART-50-mono and \textsc{SumTra} on the CrossSum Spanish and Bengali test sets.\vspace{-6pt}}
    \label{fig:paper4_catastrophic_forget_both}
\end{figure}

To this aim, Figure \ref{fig:paper4_catastrophic_forget_both} plots the relative changes in BERTScore for mBART and mBART-mono over Spanish and Bengali at an increasing number of fine-tuning samples. For this experiment we have used all the 1241 available fine-tuning samples for Bengali, and 2000 fine-tuning samples for Spanish.

For both languages, it is manifest that \textsc{SumTra} is the only model capable of a significant zero-shot performance, with a difference of approximately 30 pp compared to both mBART-50 models. At zero-shot and 10-shot, the performance of mBART-50-mono has been even lower than that of the original mBART-50, confirming the catastrophic forgetting. However, from around 100-shots, mBART-50-mono has stably overtaken mBART-50, showing that its ``forgotten'' multilingual capabilities can be restored with a sufficient amount of fine-tuning.

In the case of Spanish, mBART-50-mono has caught up with \textsc{SumTra} at 500 shots, and then progressed with a virtually identical performance. Conversely, for Bengali, both mBART-50 models have surpassed \textsc{SumTra} at 500 shots and maintained a comparable performance from there. These trends seem very interesting as they show that, while training a cross-lingual model with monolingual data undoubtedly causes a ``catastrophic forgetting'' of its multilingual capabilities at zero- and few-shots, such capabilities can be restored with a sufficient amount of fine-tuning, and even outperform an equivalent model that has not undergone monolingual training. In the case of Bengali, it also shows that a single language model can outperform our pipeline of two, most likely because it addresses the summarization and translation task in a genuinely ``joint'' manner. At the same time, it is worth noting that our pipeline can more easily and more directly take advantage of existing summarization and translation resources, as they can be independently used to train its two modules. For instance, in this case we could leverage any other En-Bn parallel corpora to boost the translator's performance. In all cases, we do not target a scenario with unlimited number of fine-tuning data; rather, a zero/few-shot one demanding minimal effort of the annotators.

\subsection{Qualitative Analysis}
\label{subsec:paper4_qualitative_analysis}

\begin{table*}[!ht]
\centering
\resizebox{\textwidth}{!}{%
% \begin{tabular}{l|l|c}
\begin{tabular}{p{0.19\textwidth} | p{0.69\textwidth} | c}
\hline
\textbf{Model} & \textbf{Summary} & \textbf{BERTScore} \\
\hline
\multirow{6}{*}{Reference} & Las autoridades estadounidenses amenazaron a la compañía tecnológica Yahoo con ponerle una multa de US\$250.000 diarios si el gigante informático no le entregaba datos de usuarios. &  \\
& \textbf{Back-Translation:} The US authorities threatened the technology company Yahoo with a daily fine of US\$250,000 if the computer giant did not provide it with user data. &  \\
\hline
mBART-50-mono (1000-shot) & \textbf{Prediction:} El gobierno de Estados Unidos publicó información sobre un caso que ha sacudido a la empresa de informática Yahoo. & \multirow{2}{*}{55.61} \\
\hline
\multirow{5}{*}{\textsc{SumTra}} & \textbf{Intermediate Summary:} The US government threatened to impose fines of up to \$250,000 (£250,000) if it refused to comply with a court order against Yahoo, according to newly released documents. & \multirow{6}{*}{61.47} \\
(100-shot) & \textbf{Prediction:} El gobierno estadounidense \textcolor{forestgreen}{\textbf{amenazaba}} con imponer multas de hasta 250.000 dólares \textcolor{red}{\textbf{(£250,000)}} \textcolor{forestgreen}{\textbf{si se niega a cumplir un}} \textcolor{blue}{\textbf{decreto judicial}} \textcolor{forestgreen}{\textbf{contra Yahoo}}, según documentos publicados recientemente. & \\
\hline
\multirow{4}{*}{\textsc{SumTra} (100-shot)} & \textbf{Intermediate Summary:} Yahoo has been fined \$250,000 (£250,000) for breaching a US government order to monitor its online services. & \multirow{5}{*}{54.78} \\
\multirow{2}{*}{(no BT loss)} & \textbf{Prediction:} Yahoo \textcolor{red}{\textbf{ha sido sancionado}} con 250.000 dólares \textcolor{red}{\textbf{(250.000 libras esterlinas)}} por \textcolor{red}{\textbf{violar un decreto del gobierno estadounidense}} para controlar sus servicios en línea. & \\
\hline
\end{tabular}
}
\caption{Qualitative example for Spanish (CrossSum). \textcolor{red}{\textbf{(Red)}} denotes incorrect translations or factual inconsistencies, \textcolor{blue}{\textbf{(Blue)}} denotes information from the source document, and \textcolor{forestgreen}{\textbf{(Green)}} refers to matching information in the reference summary.
}
\label{tab:paper4_qual_lang1}
\end{table*}

To qualitatively show that \textsc{SumTra} achieves better performance than mBART with fewer shots, in Table \ref{tab:paper4_qual_lang1} we report an example for Spanish, comparing an mBART-50-mono model fine-tuned with 1000 shots with a \textsc{SumTra} model fine-tuned with 1/10 of the shots (100). For further comparison, we also show the summary generated by \textsc{SumTra} fine-tuned without the back-translation (BT) loss of Equation \ref{equation:paper4_auxiliary_loss}. The summary generated by the mBART-50-mono model undoubtedly contains some information relevant to the reference, such as the relationship between the US authorities and Yahoo. However, it is overall generic and vague. For instance, the specific mention of a ``fine of \$250,000'' in the reference is not conveyed in the prediction. Conversely, both predictions from the \textsc{SumTra} models have been able to pick up this fact. At its turn, the prediction from the model without the BT loss has incorrectly stated that Yahoo has already been sanctioned (\textit{ha sido sancionado}), while the prediction from the full model has been in general the most informative and accurate. For example, it has been able to include the entity \textit{decreto judicial} (\textit{court order}) that is not present in the reference, but is an important piece of information in the input document (NB: Table \ref{tab:paper4_qual_lang_further_es} in Appendix \ref{subsec:paper4_appendix_qual}), and also the key term \textit{amenazaba} (\textit{threatened}). The intermediate summary in English shows that this is owed to an effective summarization, which has been carried over faithfully into the Spanish translation. However, it is also clear that the summary generated by the full \textsc{SumTra} model is still imperfect, having predicted £250,000 instead of \$250,000. Additional, commented examples are provided in Appendix \ref{subsec:paper4_appendix_qual}.

% \vspace{-6pt}

\subsection{Inference Time}
\label{subsec:paper4_gen_speed_analysis}

Given that the proposed model uses two language models in pipeline, it is important to compare its inference times to those of the baseline. To this aim, Table \ref{tab:paper4_inference_analysis} reports the inference times per sample\footnote{We have measured the inference time as the time taken to traverse the model's \texttt{generate} function, which occurs twice per sample in \textsc{SumTra} and once in mBART-50. All other overheads are negligible.} of the two models over the test sets of Spanish and Bengali. As to be expected, the proposed model has proved slower on average to generate a prediction; however, less than twice as slow: in the case of Bengali, the inference time per sample has been 1.87x that of mBART-50, and for Spanish only 1.15x. For Bengali, the larger overhead has mainly been due to an average lengthening of the predicted intermediate summaries, which has increased both the summarization and the translation times. In turn, the lengthening of the intermediate summaries has likely been induced by the back-translated summaries, which have been on average slightly longer than the references. However, the overall speed seems to have remained acceptable.

\begin{table}[!h]
\centering
\resizebox{0.75\columnwidth}{!}{%
\begin{tabular}{ccc}
\hline
\multirow{2}{*}{\textbf{Model}} & \textbf{Spanish} & \textbf{Bengali} \\ \cline{2-3}
& Per Sample (s) & Per Sample (s) \\
\hline
mBART-50 & \textbf{0.146} & \textbf{0.145} \\
\textsc{SumTra} & 0.168 & 0.271 \\
\hline
\end{tabular}
}
\caption{Average inference times per sample for mBART-50 and \textsc{SumTra} over the CrossSum Spanish and Bengali test sets.}
\label{tab:paper4_inference_analysis}
\end{table}

\section{Conclusion}
\label{sec:paper4_conclusion}

In this paper, we have proposed \textsc{SumTra}, an XLS model that revisits the traditional summarize-and-translate approach into a more contemporary end-to-end differentiable pipeline. Given that genuine XLS annotation is demanding, the main aim of the proposed model is to provide a competitive zero- and few-shot performance.

In the paper, we have evaluated the proposed approach over two mainstream XLS datasets and against a set of performing baselines, giving evidence to the competitive performance of the proposed approach. In particular, \textsc{SumTra}'s zero-shot performance has proved very strong, and its few-shot performance has been remarkable for a majority of the languages. Through various sensitivity, ablation, and qualitative analyses we have shown that the proposed model benefits from the possibility to separately train its component modules, and that its memory and inference time overheads compared to the base model are both manageable. In the future, we aim to test model configurations with different base language models (e.g., \textsc{Pisces}) for the summarization and translation modules, and explore alternative fine-tuning strategies such as adversarial training and reinforcement learning.
\vspace{18pt}

\section*{Limitations}
The proposed approach has various limitations. The most immediate is that we have limited our experimental validation to the English-to-many case. However, this was done only for the simplicity of carrying out a one-to-many set of experiments rather than a many-to-many. Instead, an actual, intrinsic limitation of the proposed approach is that it relies on a strong performance from both its summarization and translation modules. In turn, this assumes the availability of an adequate monolingual summarization training set for the source language, and an adequate parallel training corpus for the language pair---or equivalent pretrained models. However, both these requirements are much more easily met than requiring the availability of large XLS annotated resources.

The memory footprint of the proposed model, that has 1.2B total parameters, is also more imposing than that of a single, equivalent multilingual model. In particular, the memory required during fine-tuning (with the selected hyperparameters) has been approximately 34 GB. However, in Appendix \ref{subsec:paper4_specific_module_tuning} we show that it is possible to fine-tune only one of the two modules in turn (either the summarizer or the  translator) and still retain a remarkable performance, bringing back the memory requirements to those of a standard model. At its turn, the training time of the proposed model has only been approximately 1.6x times that of a single model, and should not hinder its use.

Finally, the computation of the expected embeddings in Equation \ref{equation:paper4_expected_embs} requires the product of token embeddings from the translator with the probabilities assigned to those same tokens by the summarizer. This implies that the summarizer and the translator have to share the same vocabulary, and for this reason we have built them both out of the same base model (mBART-50-large). However, it should be easy to organize a redistribution of the summarizer's probabilities over a different vocabulary, allowing mixing different base models. As a final clarification, the generation of the back-translations used for fine-tuning is conducted offline and one-off, and their auxiliary fine-tuning objective carries no measurable computational overhead.

\vspace{36pt}

\bibliography{anthology,custom}
% \bibliographystyle{acl_natbib}
% \newpage
% \phantom{newpage}
\appendix
\newpage

\section{Appendix}
\label{sec:paper4_appendix}
\subsection{Experimental Setup}
\label{subsec:paper4_appendix_expsetup}
We have selected six languages from both CrossSum and WikiLingua, and self-categorized them into high, medium, and low-resource based on the number of pretraining sentences used in \citet{tang-etal-2021-multilingual}. The groupings are selected as follows: languages with >1M pretraining sentences have been labelled as high-resource, between 100k and 1M as medium-resource, and <100K as low-resource. We refer the reader to Table 6 of \citet{tang-etal-2021-multilingual} for language-specific breakdowns.

For the evaluation of our approach, we have adopted ROUGE and BERTScore to assess both the surface and semantic matching between the predictions and the reference summaries. As mentioned in the main body, we have chosen to report the average of ROUGE-1, ROUGE-2, and ROUGE-L F1 scores, in line with previous summarisation literature. More specifically, mROUGE\footnote{\url{https://github.com/csebuetnlp/xl-sum/tree/master/multilingual_rouge_scoring}} has been used in our experiments for languages where existing language-specific stemmers and/or tokenizers are made available by the underlying package (NLTK). We note that the adoption of mROUGE in the XLS literature is not widespread, probably because its reliance on dedicated stemmers and tokenizers is somehow limiting. Given this, and a recent advocacy for BERTScore in XLS \citep{koto-etal-2021-evaluating}, we have chosen to report BERTScore extensively. To ensure that we could compute it consistently for all the languages in our evaluation, we have populated it with the weights of the encoder of the pretrained multilingual LM used for the \textsc{Tra} module of \textsc{SumTra} (\verb|mBART-large-50-one-to-many-mmt|). 

\subsection{Model Hyperparameters}
\label{subsec:paper4_appendix_modelhyp}

Our baseline model is the pretrained \verb|mBART-large-50| \citep{tang-etal-2021-multilingual}, with its variants (one-to-many\footnote{\url{https://huggingface.co/facebook/mbart-large-50-one-to-many-mmt}}, many-to-many\footnote{\url{https://huggingface.co/facebook/mbart-large-50-many-to-many-mmt}}, and many-to-one\footnote{\url{https://huggingface.co/facebook/mbart-large-50-many-to-one-mmt}}) utilized throughout the paper.
All the models have been fine-tuned and run using PyTorch Lightning on a single NVIDIA A40 GPU with 48 GB of memory. Fine-tuning the entire \textsc{SumTra} with the chosen hyperparameters uses up approximately 70\% of the total memory. Increasing the batch size and/or the input/output sequence length correspondingly increases the memory footprint, as expected. Table \ref{tab:paper4_hyperparameters} reports the full list of the hyperparameters used for training, fine-tuning  and inference.

For model training, when training the monolingual summarizer, or conducting few-shot fine-tuning of \textsc{SumTra} and the mBART-50 variants, we have selected the best checkpoints based on either a) meeting a criterion based on validation performance, or b) reaching the maximum set number of training iterations/epochs.
For mBART-50-mono, we have used the same hyperparameters as for our mBART-50 baseline model, with the exception that the former has first been trained on an English-English split of either CrossSum or WikiLingua, depending on the downstream fine-tuning dataset. This is the equivalent of the training of the \textsc{Sum} module used in \textsc{SumTra}. Lastly, for ChatGPT and davinci-003 we have used the OpenAI platform between the 18th and 28th of October 2023 (ChatGPT), and between the 12th and 28th of November 2023 (davinci-003).

\begin{table}[!t]
\centering
\resizebox{0.75\columnwidth}{!}{%
\begin{tabular}{cc}
\hline
\textbf{Hyperparameter} & \textbf{Value}\\
\hline
\multicolumn{2}{c}{\textbf{Training \textsc{Sum}}} \\
\hline
Warmup & 500 steps\\
Input Length & 512 tokens\\
Output Length & 128 tokens\\
\hline
\multicolumn{2}{c}{\textbf{Fine-Tuning \textsc{SumTra}}} \\
\hline
Warmup & 0 steps\\
Input Length & 512 tokens\\
Output Length & 84$^{\dagger}$/64$^{\ddag}$ tokens\\
Freeze Strategy & Train All \\
$\alpha$  (Eq. \ref{eq:mixed_objective}) & 0.99 \\
\hline
\multicolumn{2}{c}{\textbf{Open AI API Hyperparameters}} \\
\hline
Temperature & 0.7 \\
Frequency Penalty & 0.0 \\
Logit Bias & null \\
Log Probs & False \\
Max Tokens & 4096 -- Prompt Length \\
N & 1 \\
Presence Penalty & 0.0 \\
\hline
\multicolumn{2}{c}{\textbf{Shared Hyperparameters}} \\
\hline
Training LR & $3\times10^{-5}$ \\
Training Epochs & 10 \\
Early Stopping Criterion & 2 epochs \\
Training Batch Size & 1 \\
Inference Batch Size & 8 \\
Gradient Accumulation & 8 \\
Optimizer & AdamW \\
\hline
\end{tabular}
}
\caption{Hyperparameters used for training and evaluation of each module and the Open AI API. The ($\dagger$) and ($\ddag$) superscripts are for the CrossSum and WikiLingua datasets, respectively. With exception of Max Tokens and Temperature, the hyperparameters used with the Open AI API are default values.}
\label{tab:paper4_hyperparameters}
\end{table}

\subsection{Dataset Links and Statistics}
\label{subsec:paper4_appendix_dataset_stats}
We refer the reader to the original papers \citep{ladhak-etal-2020-wikilingua, bhattacharjee-etal-2022-crosssum} for detailed statistics of the CrossSum and WikiLingua datasets, as well as access to the original data we have made use of in this work.

For quick reference, Table \ref{tab:paper4_dataset_statistics} provides the total size of the training, validation, and test splits of the English-to-many versions of both datasets for the languages covered in our experiments. For the XSum dataset, we have downloaded the En-En data from Hugging Face. Table \ref{tab:dataset_links} provides the actual links and license types.

\begin{table}[!ht]
\centering
\resizebox{0.8\columnwidth}{!}{%
\begin{tabular}{ccccc}
\hline
\textbf{Dataset} & \textbf{Train} & \textbf{Val} & \textbf{Test} \\
\hline
\textbf{CrossSum} & 22.3K & 2.8K & 2.8K \\
\textbf{WikiLingua} & 117.4K & 16.8K & 33.5K \\
\hline
\textbf{XSum} & 204K & 11.3K & 11.3K \\
\hline
\end{tabular}
}
\caption{Total size of the training, validation and test splits for the languages covered in our experiments. For XSum, we have only used the En-En data.}
\label{tab:paper4_dataset_statistics}
\end{table}

\begin{table}[!ht]
\centering
\resizebox{\columnwidth}{!}{%
\begin{tabular}{cc}
\hline
\textbf{GitHub} & \textbf{License} \\
\hline
\url{https://github.com/csebuetnlp/CrossSum} & CC BY-NC-SA 4.0 \\
\url{https://github.com/esdurmus/Wikilingua} & CC BY-NC-SA 3.0 \\
\url{https://huggingface.co/datasets/xsum} & Unknown \\
\hline
\end{tabular}
}
\caption{GitHub repositories and license details for the CrossSum, WikiLingua, and XSum datasets.}
\label{tab:dataset_links}
\end{table}

\subsection{Fine-Tuning Ablation}
\label{subsec:paper4_specific_module_tuning}
\begin{figure}[!t]
    \centering
    \includegraphics[width=\linewidth, trim={0 0 0 0}, clip] % <- this trick for cropping was genius!
    % {different_modules.png}
    {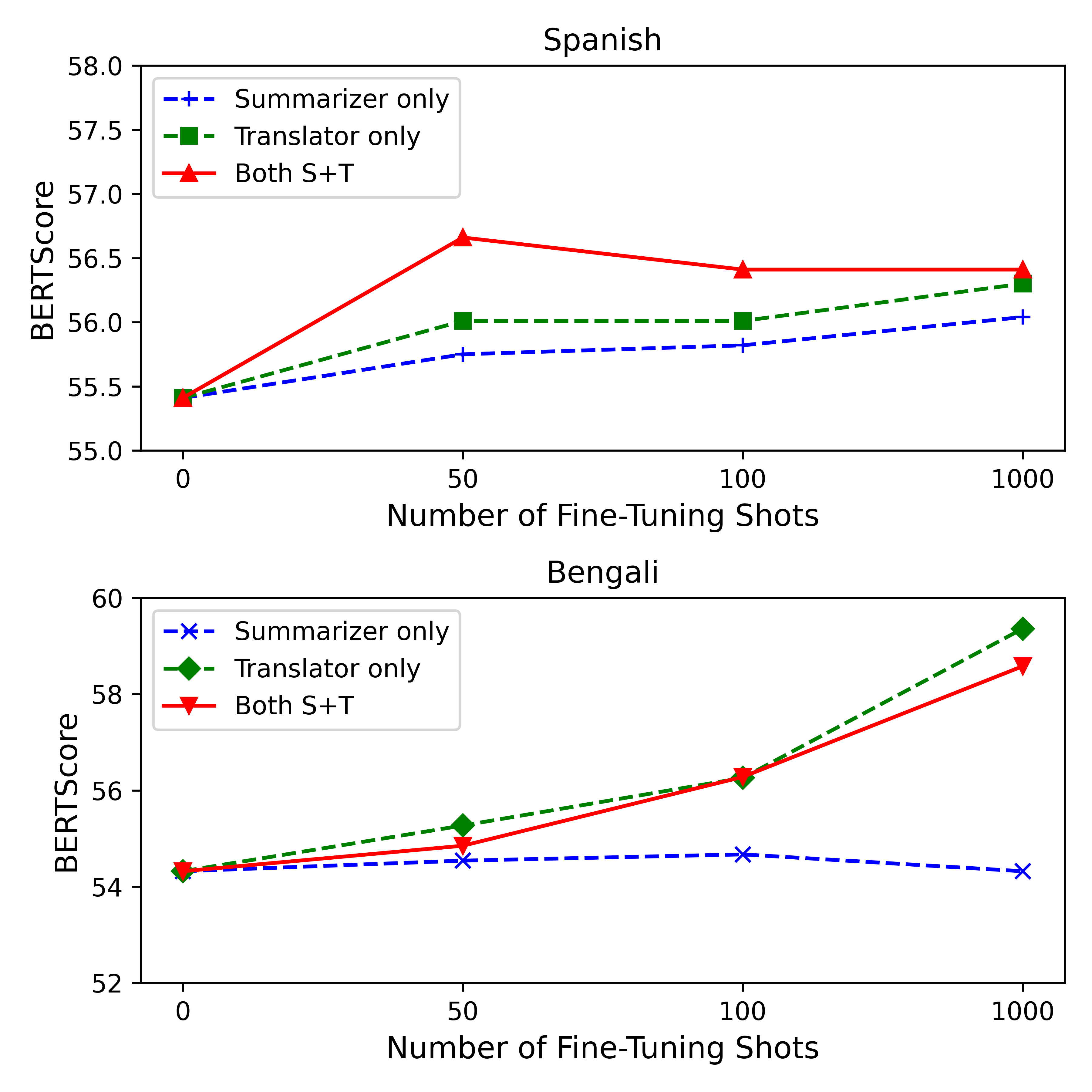}
    \caption{BERTScore scores for the CrossSum Spanish and Bengali test sets with different fine-tuning configurations (summarizer only, translator only, and both).}
    \label{fig:paper4_tuning_different_modules}
    
\end{figure}

The proposed \textsc{SumTra} model has approximately double the number of parameters of a single mBART-50-large language model. However, this is a rather small model by contemporary standards (611M parameters), and \textsc{SumTra} can comfortably fit in the memory of any standard machine for inference. Conversely, the memory footprint may become an issue for some machines in the case of fine-tuning. For this reason, we have tested \textsc{SumTra}'s performance by fine-tuning only either the summarizer or the translator, and comparing it to fine-tuning both jointly. This is to show that significant performance can still be achieved if memory constraints force the fine-tuning to be carried out at a parity of trainable parameters with mBART-50.
To this aim, Figure \ref{fig:paper4_tuning_different_modules} plots the BERTScore of the three configurations for Spanish and Bengali, with an increasing amount of fine-tuning samples. For both languages, updating only the parameters of the summarizer has led to the smallest improvements over the zero-shot performance. It could be argued that the summarizer has already been well-trained by the monolingual data, and as such its relative margin for improvement is smaller. Conversely, in the case of Bengali in particular, fine-tuning only the translator with 50 shots has achieved performance that has surpassed the tuning of both the summarizer and translator together. The trend has been the opposite for Spanish, where fine-tuning the translator alone has underperformed the fine-tuning of the entire model. This shows that the behavior of the translation component can be very language-dependent.

If memory constraints force the fine-tuning to be carried out at a parity with a single mBART-50 model, several other strategies could be easily put in place, such as alternating between updating the summarizer and the translator in turn, or fine-tuning only selected layers of the modules' encoders and decoders. However, we believe that this is not specially critical and have not explored it further.

\subsection{Sensitivity to the  Alpha Hyperparameter}
\label{subsec:paper4_appendix_alpha_weights}
% \textcolor{red}{\textit{(Can we shorten this Appendix section? Or remove?)}}
The fine-tuning objective in Equation \ref{eq:mixed_objective} combines an XLS loss and a back-translation loss with a positive coefficient, $\alpha$. The back-translation loss only influences the summarizer, while the XLS loss influences the translator directly, and the summarizer via backpropagation through the soft predictions. To explore the sensitivity of the performance to the value of the $\alpha$ coefficient, Table \ref{tab:paper4_appendix_alpha_weights} reports the mROUGE and BERTScore scores of the 100-shot \textsc{SumTra} over Spanish and Bengali for increasing $\alpha$ values (i.e., increasing relative influence of the back-translation loss).

The results show that in the case of Spanish the best $\alpha$ value has been rather high (0.95), likely because the pretrained translator is already good enough for this language, and the emphasis has been on keeping the summarization aligned with the target. Conversely, in the case of Bengali the relative weight of the XLS loss for the best performance has been much higher (0.50), showing that for this lower-resource language the updates to the translator have proved more important.

For our experiments, we could have grid-searched an optimal value of $\alpha$ for every language---which would have made our model perform even better---or just use a trade-off value for all languages, which is more practical and convenient for prospective users. In the interest of usability, we have chosen to not over-validate $\alpha$, selecting a somehow arbitrary fixed value of 0.99 to emphasize the back-translation loss in all cases.

\begin{table}[!ht]
\centering
\resizebox{0.65\columnwidth}{!}{%
\begin{tabular}{ccc}
\hline
$\alpha$ & \textbf{Spanish} & \textbf{Bengali} \\
\hline
0.00 & 21.04 / 56.44 & 4.20	/ 55.54 \\
0.50 & 20.76 / 56.20 & \textbf{5.21}	/ \textbf{56.38} \\
0.90 & 21.30 / 56.46 & 4.58 / 56.02 \\
0.95 & \textbf{21.43} / \textbf{56.56} & 4.25 / 55.65 \\
0.99 & 21.37 / 56.41 & 4.67 / 56.28 \\
1.00 & 19.96 / 55.33 & 3.81	/ 54.61 \\
\hline
\end{tabular}
}
\caption{mROUGE and BERTScore scores for different $\alpha$ values in the objective function (CrossSum).}
\label{tab:paper4_appendix_alpha_weights}
\end{table}

% It is perhaps worth remarking once more the respective impact of these two losses on the summarizer: the back-translation loss keeps the predicted summaries more closely aligned with the references, while the XLS loss only influences the summarizer in a ``looser'' and indirect way via backpropagation of the translator's gradients. Therefore, removing the back-translation loss altogether, or likewise keeping $\alpha$ very low, leads to summaries that still seem qualitatively very effective, but are less faithful to the target. This tends to penalize the scores, especially mROUGE, but did not seem undesirable to us from a qualitative perspective. We leave further exploration and evaluation of this trade-off to future work.

% In contrast, our model tuned on Spanish is less receptive when only tuning the translator and instead performs best when both modules are tuned in the pipeline.

% It should be noted that in the same vein of the work presented by \citet{unanue-etal-2023-t3l}, this network architecture facilitates the ability to selectively tune not just the \textsc{Sum} or \textsc{Tra} modules individually, but particular layers in the encoder and/or decoder of each.

\subsection{Sensitivity to Different Embedding-based Metrics}
\label{subsec:paper4_appendix_moverscore}
As a further sensitivity analysis, we explore the sensitivity of the results to the BERTScore evaluation metric by comparing it with MoverScore \citep{zhao-etal-2019-moverscore}. These two metrics are rather similar, as they are both variants of optimal transport. However, their main difference is that BERTScore performs a one-to-one alignment between the tokens of the prediction and the reference, while MoverScore performs a one-to-many, allowing a token to receive a good matching score from the accumulation of multiple, partial matches\footnote{For computing MoverScore, we have used BERT-base-multilingual-uncased (\url{https://huggingface.co/bert-base-multilingual-uncased}).}.

\begin{figure}[!t]
    \centering
    \includegraphics[width=\linewidth, trim={0 0 0 0}, clip]{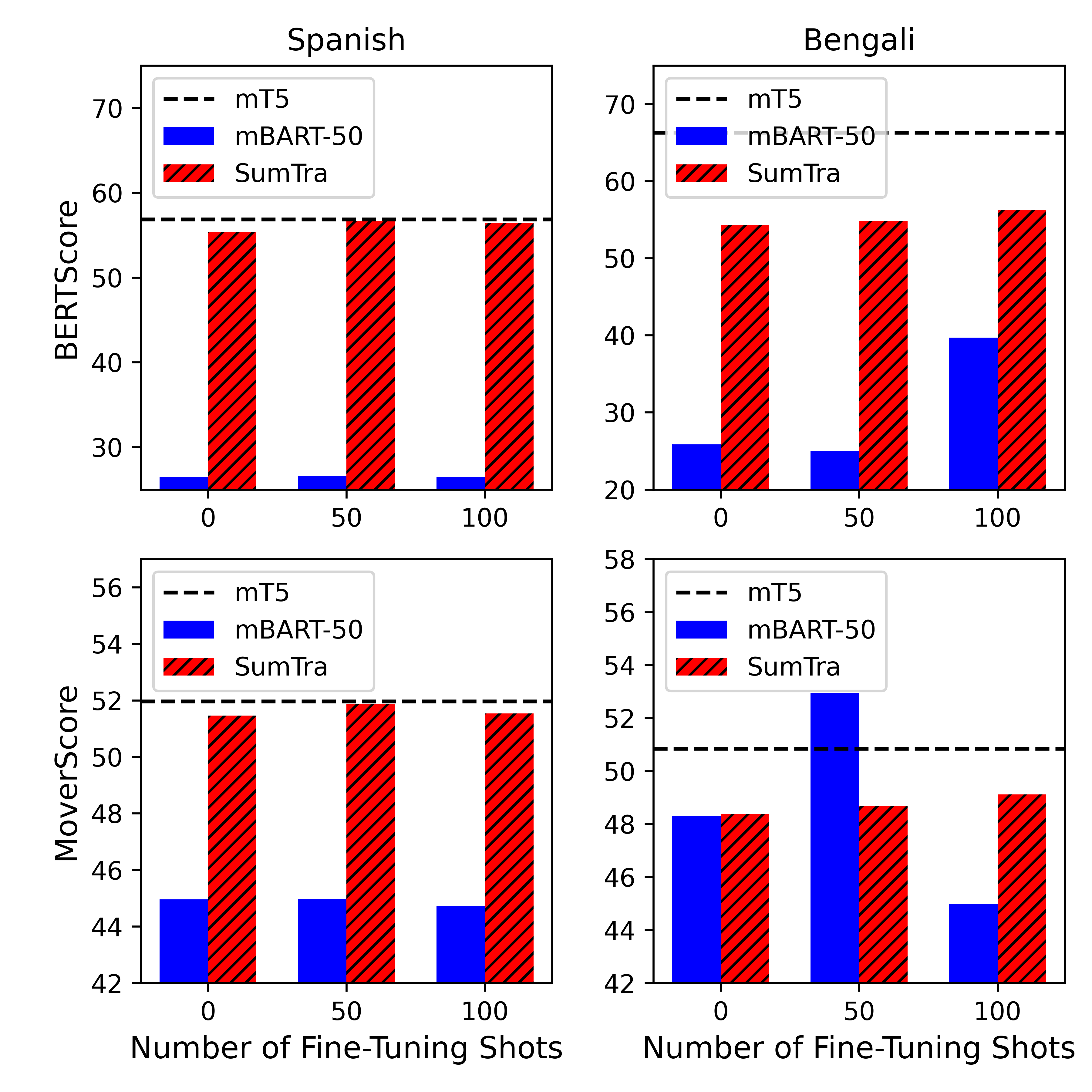}
    \caption{BERTScore and MoverScore comparison over the Spanish and Bengali test sets (CrossSum)).}
    \label{fig:paper4_appendix_moverscore}
\end{figure}

Figure \ref{fig:paper4_appendix_moverscore} shows the BERTScore and MoverScore values for mBART-50 and SumTra for Spanish and Bengali in zero- and few-shot configurations. In addition, the values for the fully-trained mT5(m2m) are displayed for reference as an informal upper-bound. For Spanish, the qualitative trends for BERTScore and MoverScore are similar, with the only notable difference that the MoverScore values are more compressed in range. For Bengali, the trends have instead differed significantly, with the MoverScore values for mBART-50 and \textsc{SumTra} being roughly on par on average. However, the MoverScore results for mBART-50 show a very marked drop for 100-shot fine-tuning, which seems to contradict the qualitative evaluation and the expected impact from fine-tuning. For this reason, we have chosen to report BERTScore in the main paper.

\subsection{Soft vs. Hard Predictions at Inference Time}
\label{subsec:paper4_appendixsoftvhard}
In the proposed model, the use of soft predictions is strictly required during fine-tuning, but becomes an option at inference time. For this reason, in this section, we examine the impact of using either soft or hard predictions for inference. As hard predictions, we simply extract the argmaxed predictions from the summarizer and pass them to the translator, without converting them to embedding space and bypassing the embedding layer of the translator. 

% Table \ref{tab:paper4_appendixsoftvhard_result} shows the results over the CrossSum Spanish test set for a 100-shot configuration for both cases. It is clear that the hard predictions have led to noticeably better scores. While this is only for a single language, it is reasonable to assume that these results may generalize to other languages, given that using the argmax provides a more confident and tighter input to the translation module.

To showcase the differences, Table \ref{tab:paper4_appendixsoftvhard_result} presents a short qualitative example. For both types of predictions, we have fine-tuned the model using the soft predictions, but passed either hard or soft predictions to the translator module for inference. For clarity, the summarizer generates the same intermediate summary in both cases. As the BERTScore values show, there is little semantical difference between the two types of prediction. However, given that the argmax has obtained a mildly higher score (alongside a minor inference speedup), we have chosen to use the hard predictions throughout our experiments. While these results are only for a single language, it is reasonable to assume that they may generalize to other languages, given that using the argmax provides a more confident and tighter input to the translation module.

% \begin{table}[!h]
% \centering
% \resizebox{0.9\columnwidth}{!}{%
% \begin{tabular}{ccc}
% \hline
% \textbf{Type} & \textbf{BERTScore} & \textbf{MoverScore} \\
% \hline
% \multicolumn{3}{c}{\textbf{Spanish}} \\
% \hline
% mT5-m2m & \textit{56.86} & \textit{51.96} \\
% mBART-50 (0-shot) & 26.46 & 44.95 \\
% mBART-50 (50-shot) & 26.54 & 44.98 \\
% mBART-50 (100-shot) & 26.50 & 44.73 \\
% \hline
% \textsc{SumTra} (0-shot) & 55.41 & 51.46 \\
% \textsc{SumTra} (50-shot) & \textbf{56.66} & \textbf{51.87} \\
% \textsc{SumTra} (100-shot) & 56.41 & 51.54 \\
% \hline
% % \hline
% \multicolumn{3}{c}{\textbf{Bengali}} \\
% \hline
% mT5-m2m & \textit{66.31} & 50.84 \\
% mBART-50 (0-shot) & 25.83 & 48.31 \\
% mBART-50 (50-shot) & 25.00 & \textit{52.95} \\
% mBART-50 (100-shot) & 39.70 & 44.98 \\
% \hline
% \textsc{SumTra} (0-shot) & 54.32 & 48.37 \\
% \textsc{SumTra} (50-shot) & 54.85 & 48.67 \\
% \textsc{SumTra} (100-shot) & \textbf{56.28} & \textbf{49.11} \\
% \hline
% \end{tabular}
% }
% \caption{MoverScore results for Spanish and Bengali test sets (CrossSum). The best scores are boldfac ed.}
% \label{tab:paper4_appendix_moverscore}
% \end{table}

\subsection{Additional Qualitative Analysis}
\label{subsec:paper4_appendix_qual}

\begin{table*}[!thb]
\centering
\resizebox{0.99\textwidth}{!}{%
\begin{tabular}{p{0.22\textwidth} | p{0.65\textwidth} | c}
\hline
\textbf{Model} & \textbf{Summary} & \textbf{BERTScore} \\
\hline
\multirow{3}{*}{Reference} & Un hombre demasiado asustado para volar debido a la pandemia vivió sin ser detectado en un área segura del aeropuerto internacional de Chicago durante tres meses, según los fiscales de EE.UU. & \\
\hline
\multirow{3}{*}{Intermediate Summary} & A man arrested after allegedly stealing a badge from an airport in Chicago was "unauthorised, non-employee" according to the official prosecutor. & \\
\hline
\multirow{3}{*}{Argmax} & \textbf{Prediction:} Un hombre detenido después de haber supuesto \textcolor{blue}{\textbf{robo de un badge}} en un \textcolor{forestgreen}{\textbf{aeropuerto de Chicago}} fue "\textcolor{blue}{\textbf{no autorizado}}, \textcolor{red}{\textbf{no asalariado}}" según el fiscal oficial. &  \multirow{3}{*}{56.03} \\
\hline
\multirow{3}{*}{Soft} & \textbf{Prediction:} Un hombre detenido por supuesto \textcolor{red}{\textbf{robo de un cohete}} de un \textcolor{forestgreen}{\textbf{aeropuerto de Chicago}} fue "\textcolor{blue}{\textbf{no autorizado}}", \textcolor{blue}{\textbf{no trabajador}}", según el fiscal oficial. &  \multirow{3}{*}{55.43} \\
\hline
\end{tabular}
}
\caption{Qualitative example to support the use of the hard vs. soft predictions at inference time (CrossSum Spanish). \textcolor{red}{\textbf{(Red)}} denotes incorrect translations or factual inconsistencies, \textcolor{blue}{\textbf{(Blue)}} denotes information from the source document, and \textcolor{forestgreen}{\textbf{(Green)}} refers to matching information in the reference summary.}
\label{tab:paper4_appendixsoftvhard_result}
\end{table*}

\begin{table*}[!ht]
\centering
\resizebox{0.99\textwidth}{!}{%
% \begin{tabular}{l|l|c}
\begin{tabular}{p{0.20\textwidth} | p{0.65\textwidth} | c}
\hline
\textbf{Model} & \textbf{Summary} & \textbf{BERTScore} \\
\hline
\multirow{4}{*}{Reference} & Buatlah sayap. Buatlah lingkaran cahaya. Kombinasikan sayap dan lingkaran cahaya dengan kostum. &  \\
& \textbf{Back-Translation:} Make wings. Make circles of light. Combine wings and circles of light with costumes. & \\ 
\hline
\multirow{4}{*}{\textsc{SumTra} (100-shot)} & 
\textbf{Intermediate Summary:} Make or buy wings. Make or buy a halo. Make or buy a scarf. & \multirow{4}{*}{57.54} \\
& \textbf{Prediction:} \textcolor{forestgreen}{\textbf{Buat}} atau beli \textcolor{forestgreen}{\textbf{sayap}}. \textcolor{forestgreen}{\textbf{Buat}} atau beli \textcolor{blue}{\textbf{halo}}. \textcolor{forestgreen}{\textbf{Buat}} atau beli \textcolor{red}{\textbf{kain jambu.}} &  \\
\hline
\multirow{3}{*}{\textsc{SumTra} (100-shot)} & \textbf{Intermediate Summary:} Angel wings are a way of decorating your Halloween costume. & \multirow{4}{*}{45.63}\\
(no BT loss) & \textbf{Prediction:} \textcolor{red}{\textbf{Burung-burung}} \textcolor{blue}{\textbf{malaikat}} adalah \textcolor{red}{\textbf{cara untuk mengecatkan}} kostum \textcolor{red}{\textbf{Halloween Anda}}. & \\
\hline
\end{tabular}
}
\caption{Qualitative example for Indonesian (WikiLingua) for \textsc{SumTra} (100-shot) with and without the use of the back-translation (BT) loss. \textcolor{red}{\textbf{(Red)}} denotes incorrect translations or factual inconsistencies, \textcolor{blue}{\textbf{(Blue)}} denotes information from the source document, and \textcolor{forestgreen}{\textbf{(Green)}} refers to matching information in the reference summary.}
\label{tab:paper4_qual_lang2}
\end{table*}

\begin{table*}[!ht]
\centering
\resizebox{\textwidth}{!}{%
% \begin{tabular}{l|l|c}
\begin{tabular}{p{0.19\textwidth} | p{0.69\textwidth} | c}
\hline
\textbf{Model} & \textbf{Summary} & \textbf{BERTScore} \\
\hline
\multirow{17}{*}{Input Document} & According to court documents, the National Security Agency (NSA) had demanded that Yahoo comply with new surveillance rules, something the company said was unconstitutional. Yahoo failed in a court challenge on the constitutionality of the order. But the details emerged on Thursday when a federal judge ordered the unsealing of some material about the case. Yahoo's general counsel Ron Bell said publication of the material was "an important win for transparency". Yahoo said that the government amended a law to demand user information from online services, prompting a court challenge. Former NSA contractor Edward Snowden disclosed the programme last year. But the court documents reveal that the battle over surveillance between technology firms and the US government stretched back years before the Snowden revelations. The new material about the case, first reported by the Washington Post, underscores "how we had to fight every step of the way to challenge the US government's surveillance efforts',' Mr Bell added. "At one point, the US government threatened the imposition of \$250,000 in fines per day if we refused to comply," he said. About 1,500 pages of previously classified documents were unsealed by a federal court. & \\
\hline
\multirow{3}{*}{Reference} & Las autoridades estadounidenses amenazaron a la compañía tecnológica Yahoo con ponerle una multa de US\$250.000 diarios si el gigante informático no le entregaba datos de usuarios. &  \\
\hline
\multirow{3}{*}{\textsc{SumTra} (100-shot)} & El gobierno estadounidense amenazaba con imponer multas de hasta 250.000 dólares (£250,000) si se niega a cumplir un decreto judicial contra Yahoo, según documentos publicados recientemente. & \multirow{3}{*}{61.47} \\
\hline
\multirow{19}{*}{davinci-003 (ST)} & Según los documentos judiciales, la Agencia de Seguridad Nacional (NSA) exigió que Yahoo cumpliera con nuevas reglas de vigilancia, algo que la empresa dijo que era inconstitucional. Yahoo fracasó en un desafío judicial sobre la constitucionalidad de la orden. Pero los detalles surgieron el jueves cuando un juez federal ordenó desclasificar algunos materiales sobre el caso. El abogado general de Yahoo, Ron Bell, dijo que la publicación de los materiales fue "una importante victoria para la transparencia". Yahoo dijo que el gobierno modificó una ley para exigir información de los usuarios de los servicios en línea, lo que desencadenó un desafío judicial. El contratista de la NSA Edward Snowden dio a conocer el programa el año pasado. Pero los documentos judiciales revelan que la batalla entre las empresas tecnológicas y el gobierno de Estados Unidos se extendió durante años antes de las revelaciones de Snowden. Los nuevos materiales sobre el caso, informados por primera vez por el Washington Post, subrayan "cómo tuvimos que luchar en cada paso para desafiar los esfuerzos de vigilancia del gobierno de EE. UU.'', Agregó el Sr. Bell. "En un momento, el gobierno de EE. amenazó con la imposición de multas de \$250,000 por día si nos negábamos a cumplir", dijo. Unos 1.500 páginas de documentos previamente clasificados fueron desclasificados por un tribunal federal. & \multirow{19}{*}{48.15} \\
\hline
\multirow{5}{*}{ChatGPT (Direct)} & De acuerdo con documentos judiciales, la Agencia de Seguridad Nacional (NSA) exigió que Yahoo cumpliera con nuevas reglas de vigilancia, algo que la compañía consideró inconstitucional. Yahoo perdió en un desafío legal sobre la constitucionalidad de la orden. Sin embargo, los detalles salieron a la luz el jueves cuando un j & \multirow{5}{*}{47.96} \\
\hline
\end{tabular}
}
\caption{Further qualitative example for Spanish (CrossSum) comparing the proposed approach with large language models, namely ChatGPT and davinci-003. We have included the input document here for reference.}
\label{tab:paper4_qual_lang_further_es}
\end{table*}

To supplement Table \ref{tab:paper4_qual_lang1}, in Table \ref{tab:paper4_qual_lang2} we show another qualitative example from WikiLingua for Indonesian. For this example, we have only compared \textsc{SumTra} with and without the use of the back-translation loss.
Without the back-translation loss, the summary predicted by \textsc{SumTra} has made reference to angel birds (\textit{burung-burung malaikat}) and painting (\textit{cara untuk mengecatkan}) as a means of decorating a costume. The prediction has also included an incorrect capitalization of ``you'' (\textit{Anda}). While we can roughly infer what the predicted summary means, the summary predicted by \textsc{SumTra} with the back-translation loss has made the conveyed meaning much clearer. Specifically, \textsc{SumTra} with the back-translation loss has referred to making wings (\textit{buat sayap}) and a halo (\textit{halo}), aligning more closely with the meaning of the reference summary (e.g., \textit{buatlah sayap}). Like in the qualitative example in Table \ref{tab:paper4_qual_lang1}, even this summary is still imperfect, as we note a false generation of the phrase ``\textit{kain jambu}''. However, as mentioned in the main paper, we expect that for low-resource languages such as Indonesian, a dedicated training of the translator should be able to improve the translation quality and further boost BERTScores.

Additionally, to qualitative assess the performance of ChatGPT and davinci-003, Table \ref{tab:paper4_qual_lang_further_es} shows their predictions for the same example displayed in Table \ref{tab:paper4_qual_lang1} in the main paper. In the case of davinci-003, the summarize-then-translate prompt has not worked very well in terms of length reduction, since the generated output has still come out relatively long. However, details of the input document have been relayed well in the generated summary. In contrast, the direct prompt used with ChatGPT has been effective at generating a shorter summary. However, the summary is truncated and has modest semantic correlation with the reference, as reflected by its low BERTScore. In contrast, the 100-shot \textsc{SumTra} model has retained a higher alignment with the reference summary (+10 pp BERTScore). As stated in Section \ref{sec:paper4_results}, these two LLMs have not been able to match the task-specific capability of the dedicated, smaller models (mBART-50, \textsc{SumTra}, \textsc{Pisces}).

\end{document}